\documentclass[letterpaper, 10pt, conference]{ieeeconf}

\IEEEoverridecommandlockouts                              

\usepackage[margin=54pt]{geometry}

\usepackage{silence}
\usepackage{fancyhdr}
\fancypagestyle{withfooter}{
  
  \fancyhead[L]{}
  \fancyhead[R]{}
  \fancyfoot[C]{\footnotesize Presented at the 2025 IEEE ICRA Workshop on Field Robotics}
}

\usepackage[l2tabu,orthodox]{nag}

\usepackage[
    backend=biber,
    style=ieee,
    sorting=none,
    natbib=true,
    doi=false,
    isbn=false,
    url=false,
    eprint=false,
    maxcitenames=1,
    mincitenames=1
]{biblatex}

\usepackage[pdftex,colorlinks]{hyperref}

\usepackage[printonlyused]{acronym}

\usepackage[all]{nowidow}

\usepackage[dvipsnames]{xcolor}

\usepackage{lipsum}

\usepackage{xspace} 

\setlipsum{par-before = \begingroup\color{gray}, par-after = \endgroup}

\usepackage[pdftex]{graphicx}

\usepackage{epstopdf}

\usepackage{import}

\graphicspath{{./latexGoodPractices/}}

\usepackage{booktabs}

\usepackage{tabularx}
\usepackage{multirow, multicol}

\usepackage{siunitx}
\definecolor{linkblue}{HTML}{14406E}
\hypersetup{
  colorlinks = true,
  linkcolor  = linkblue,   
  citecolor  = linkblue,   
  urlcolor   = linkblue,
  filecolor  = linkblue,
}

\DeclareSourcemap{
  \maps[datatype=bibtex]{
    \map[overwrite]{
      \step[fieldsource=journal, match=\regexp{Proceedings\s+of\s+the\s+Institution\s+of\s+Mechanical\s+Engineers.*}, replace={Proc. Inst. Mech. Eng. D}]
      \step[fieldsource=journal, match=\regexp{IEEE\s+Transactions\s+on\s+Robotics\s+and\s+Automation}, replace={IEEE Trans. Robot. Autom.}]
      \step[fieldsource=journal, match=\regexp{IEEE\s+Transactions\s+on\s+Robotics}, replace={IEEE Trans. Robot.}]
      \step[fieldsource=journal, match=\regexp{IEEE\s+Transactions\s+on\s+Aerospace\s+and\s+Electronic\s+Systems}, replace={IEEE Trans. Aerosp. Electron. Syst.}]
      \step[fieldsource=journal, match=\regexp{IEEE\s+Transactions\s+on\s+Vehicular\s+Technology}, replace={IEEE Trans. Veh. Technol.}]
      \step[fieldsource=journal, match=\regexp{IEEE\s+Robotics\s+and\s+Automation\s+Letters}, replace={IEEE Robot. Autom. Lett.}]
      \step[fieldsource=journal, match=\regexp{IEEE/ASME\s+Transactions\s+on\s+Mechatronics}, replace={IEEE/ASME Trans. Mechatron.}]
      \step[fieldsource=journal, match=\regexp{Journal\s+of\s+Intelligent.*Robotic\s+Systems}, replace={J. Intell. Robot. Syst.}]
      \step[fieldsource=journal, match=\regexp{Journal\s+of\s+Dynamic\s+Systems,\s+Measurement,\s+and\s+Control}, replace={J. Dyn. Syst. Meas. Control}]
      \step[fieldsource=journal, match=\regexp{Journal\s+of\s+Field\s+Robotics}, replace={J. Field Robot.}]
      \step[fieldsource=journal, match=\regexp{Journal\s+of\s+Terramechanics}, replace={J. Terramech.}]
      \step[fieldsource=journal, match=\regexp{Robotics\s+and\s+Autonomous\s+Systems}, replace={Robot. Auton. Syst.}]
      \step[fieldsource=journal, match=\regexp{Computers\s+and\s+Electronics\s+in\s+Agriculture}, replace={Comput. Electron. Agric.}]
      \step[fieldsource=journal, match=\regexp{Applied\s+Ocean\s+Research}, replace={Appl. Ocean Res.}]
      \step[fieldsource=journal, match=\regexp{SAE\s+International\s+Journal\s+of\s+Passenger\s+Cars.*}, replace={SAE Int. J. Passeng. Cars}]
      \step[fieldsource=journal, match=\regexp{Mechanical\s+Systems\s+and\s+Signal\s+Processing}, replace={Mech. Syst. Signal Process.}]
      \step[fieldsource=booktitle, match=\regexp{.*Intelligent\s+Robots.*Systems.*}, replace={Proc. IEEE/RSJ IROS}]
      \step[fieldsource=booktitle, match=\regexp{.*Robotics\s+and\s+Automation.*ICRA.*}, replace={Proc. IEEE ICRA}]
      \step[fieldsource=booktitle, match=\regexp{.*International\s+Conference\s+on\s+Robotics\s+and\s+Automation.*}, replace={Proc. IEEE ICRA}]
    }
  }
}

\usepackage{amssymb,amsfonts,amsmath,amscd}

\usepackage{bm}

\usepackage[capitalize,, french,english,noabbrev,nameinlink]{cleveref}
\crefname{figure}{Fig.}{Figs.}
\Crefname{figure}{Fig.}{Figs.}
\crefname{equation}{Eq.}{Eqs.}
\Crefname{equation}{Eq.}{Eqs.}

\newcommand{\bbm}{\begin{bmatrix}}
\newcommand{\ebm}{\end{bmatrix}}

\usepackage{mathptmx} 

\acrodef{MPC}[MPC]{Model Predictive Control}
\acrodef{IMU}[IMU]{Inertial Measurement Unit}
\acrodef{DOF}[DOF]{Degree of Freedom}
\acrodef{ICP}[ICP]{Iterative Closest Point}
\acrodef{RPE}[RPE]{Relative Pose Error}
\acrodef{COM}[COM]{Center of Mass}
\acrodef{ICR}[ICR]{Instantaneous Center of Rotation}
\acrodef{MTT-154}[MTT-154]{Martel Transportation Technology 154}
\acrodef{IQR}[IQR]{interquartile range}

\title{\LARGE \bfseries
Inertial Steering and Motion Prediction of Tracked Vehicles on Low-Adherence Surfaces
}

\author{
Mathis Turgeon-Roy\textsuperscript{1},
Mohamed Ounally\textsuperscript{1}\textsuperscript{2},
Veronica Vannini\textsuperscript{1},
Olivier Roy\textsuperscript{1},
François Pomerleau\textsuperscript{1}
\\
\textsuperscript{1}\textit{Northern Robotics Laboratory (Norlab), Université Laval},
Québec, QC, Canada
\\
\textsuperscript{2}\textit{TSCF, INRAE},
Clermont-Ferrand, France
}

\begin{document}

\maketitle
\thispagestyle{empty}
\pagestyle{empty}

\begin{abstract}

Tracked vehicles are widely used in off-road applications, but their conventional motion models often rely on no-slip assumptions that become inaccurate on low-adherence surfaces. 
Rather than treating inertial effects arising from slip as a disturbance to be minimized, this work investigates how inertial effects arising from a moving mass can be deliberately exploited for steering on low-friction surfaces. 
A dynamic model for an articulated tracked vehicle with a passive trailer is developed, allowing inertial effects from trailer motion to be captured. 
The approach is experimentally evaluated on ice, where our dynamic model outperforms the kinematic baseline. 
We also propose a mixed dynamic-kinematic model to adapt predictions to both surface and situation by exploiting predictions from both methods. 
Finally, we demonstrate inertial steering on a mono-track vehicle without a conventional steering mechanism, showing that controlled movements of an internal mass can generate turning motion on ice. 
The resulting behavior can be predicted using the proposed dynamic model by representing the internal mass as a frictionless trailer, revealing the analogous dynamics between the two configurations. 
These results demonstrate the feasibility of inertial steering as an alternative steering mechanism for tracked vehicles operating in low-adherence environments.

\end{abstract}


\section{INTRODUCTION}
\begin{figure}
   \centering
   \includegraphics[width=1\linewidth]{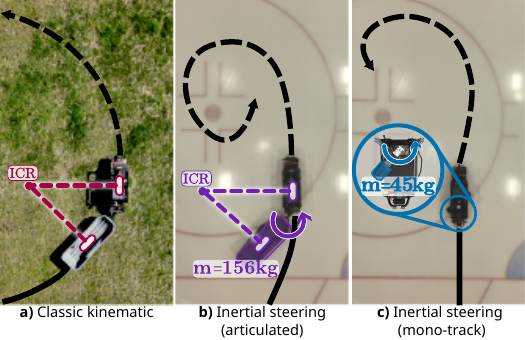}
   \caption{Motion of the MTT-154 on high- and low-adherence ground.
   (a)~On grass, the articulated vehicle follows the classic kinematic model \ac{ICR} computation, which depends only on the track speed and the articulation angle.
   (b)~On ice, the angular acceleration of the trailer mass becomes relevant to the motion of the articulated vehicle. 
   (c)~The same inertial effect steers a mono-track, which has no steering mechanism, through an actuated internal mass. 
   Low friction degrades kinematic models but opens the opportunity for a steering mechanism based on inertia.}
   \label{fig:figure1}
   \vspace{-4pt}
 \end{figure}

Tracked vehicles are a well-established design for off-road heavy-duty tasks in variable ground conditions due to their augmented surface of contact~\citep{tejiSurveyOffRoadMobile2023a}.
Their popularity makes motion modeling of tracked vehicles an important concern for autonomous systems acting in domains such as agriculture~\citep{febbraroObjectorientedModellingTracked2025}, mining~\citep{wangInvestigationTrackedMining2024}, and defense~\citep{wongEvaluationComputerSimulation2019}. 
While skid steer remains the most common track configuration, tracked vehicle designs have diversified~\citep{bruzzoneTrackedLocomotionSystems2022}, and articulated vehicles that steer by reorienting a passive trailer relative to the towing body are one example, as pictured in \cref{fig:figure1}~a).
Traditionally, motion models for tracked vehicles assume perfect adherence to the ground at a point of the track and determine the \ac{ICR} accordingly through simple kinematic models~\citep{hongVerificationWheeledMobile1999}.
Nonetheless, on low-adherence surfaces such as ice, the ability to generate lateral forces diminishes, and the accuracy of simple kinematic models is severely hindered~\citep{balakrishnaModelingSlipWheeled1995}.

In the past, slip has been mainly treated as an unknown disturbance to be estimated, compensated, or minimized, rather than as a locomotion mechanism to be deliberately exploited~\citep{gonzalezSlippageEstimationCompensation2018}.
One estimation approach augments classical kinematic models with slip parameters identified either online or from prior terrain characterization~\citep{wardDynamicModelBasedWheelSlip2008,gustafssonSlipbasedTireroadFriction1997}.
In control publications, these models are commonly integrated into \ac{MPC} to regulate slip as ground adherence changes~\citep{singhIntelligentTireBased2013}. 
The resulting behavior is safe but conservative, as the vehicle slows down precisely where friction saturates, limiting potential mobility.
Yet for certain robots, losing ground adherence also opens the opportunity for an alternative steering mechanism, with articulated vehicles serving as a representative example. 
As illustrated in \cref{fig:figure1}~b), when reorienting its trailer, an articulated vehicle also moves a mass $m$ relative to its towing body, which generates an inertial steering wrench.
On ice, as the kinematic model loses fidelity, the trailer can accelerate and move faster, increasing its inertial contribution to steering.
Inertial steering then appears as a complementary mechanism to maintain or even increase mobility on low-friction surfaces.

In this paper, we investigate the mixed use of kinematic and dynamic motion models for an articulated tracked vehicle with a passive trailer to reduce prediction error in adverse traction conditions.
Moreover, we pushed the concept of dynamic modeling further by predicting the motion of a vehicle without classical steering. 
As depicted in \cref{fig:figure1}~c), we implemented a mass-shifting system on a stand-alone track that we will refer to as a mono-track configuration in this text.

More specifically, the contributions of this paper are:
\begin{enumerate}
    \item An adaptation of a modular contact friction model for articulated track and mono-track vehicles on ice;
    \item A derivation and an experimental validation of a mixed inertial and friction steering model for an articulated vehicle; and
    \item An experimental validation of a mono-track vehicle steered exclusively through intentional inertial variation.
\end{enumerate}

\section{Related work}

\subsection{Track modeling}
On rigid ground, track models usually treat friction as the only horizontal interaction force, neglecting bulldozing and track sinkage~\citep{wong2022theory}.
\citet{wongGeneralTheorySkid2001a} laid the basis of track-to-rigid-ground interaction by modeling shear stress as a function of shear displacement.
Then, they applied the model to a skid-steer tracked vehicle, integrating the shear stress over the track contact surface to obtain the total friction wrench during steady-state operation~\citep{wongGeneralTheorySkid2001a}.
\citet{ozdemirNewContactSlip2017} extended this theory to transient states, improving predictions at near-zero track speed.
Later variants of these models reduced computation time~\citep{zhangTerramechanicsbasedDynamicModel2024} or added factors such as air resistance at high speed~\citep{milicevicTheoreticalModelHighspeed2023}.
Meanwhile, other works suggest that wheel models can predict the track behavior well~\citep{mahalingamNovelAlternateMultibody2021}, and \citet{zhangSimilaritiesWheelsTracks2024} used a wheel model to approximate the dynamics of a steel tracked skid-steer vehicle.
Similarly, \citet{maclaurinSkidSteeringModel2011b} developed a skid-steer model computing the interaction of the rubber track and the ground with a tire model.
One limitation of the research on tracked modeling is that it has heavily focused on skid-steer track configurations, and very little has been done to assess track dynamics under different configurations.
Given the diversity of vehicle track configurations in robotics~\citep{bruzzoneTrackedLocomotionSystems2022}, there is a need to evaluate track models outside of the skid-steer configuration.

\subsection{Articulated vehicle model}
Early work on articulated vehicles has been motivated by the gain in maneuverability obtained by connecting two skid-steer vehicles in series through a passive joint~\citep{alhimdaniSteeringAnalysisArticulated1982}.
Articulated vehicles with a redundantly actuated joint have also been modeled~\citep{totaArticulatedSteeringControl2021}, but this literature usually treats steering as coming from differential drive of the track.
For the tractor-trailer articulated configuration, motion models in the literature pursue different objectives with different methods.
Slip estimation was incorporated into the kinematics of a vehicle with a passive trailer for off-road trailer path following by \citet{cariouPathFollowingVehicletrailer2010}. 
Similarly, \citet{mollKinematicDynamicModeling2025} considered trailer path following and derived models for multiple trailer configurations, including actuated articulation.
Trailers with actuated wheel orientations~\citep{islamClosedloopDynamicSimulationbased2012} or braking~\citep{hac2008stability} have similarly been studied to improve handling performance and reduce energy consumption in road scenarios.
The addition of an actuated hitch to reduce off-tracking of the trailer has also been considered for road vehicles with trailer attachments~\citep{percyNumericalControlAlgorithm2012}.
Although actuated articulations typically assist trailer path following, they can also generate steering yaw for an unsteered towing body, effectively reducing the total system actuator count.
To the best of our knowledge, the motion of such a system has not been modeled yet.

\subsection{\texorpdfstring{\ac{COM}}{COM} control and inertial steering in wheeled and track robots}
\ac{COM} control has been extensively investigated within mobile wheeled and tracked robotics to enhance platform stability and traction.
For example, early foundational work by \citet{iagnemmaControlRoboticVehicles2003} introduced a control algorithm for articulated suspension rovers, demonstrating that dynamic \ac{COM} shifting relative to terrain contact points mitigates tip-over risk on rough terrain.
Building upon this framework, \citet{nakamuraWheeledRobotMovable2007} established that active inertial positioning not only preserves tip-over stability but also optimizes traction distribution across ground contact interfaces.
Expanding these principles to dynamic regimes, \citet{zhangSteeringControlHighSpeed2025} leveraged the articulated degrees of freedom in quadrupedal wheeled systems to stabilize high-speed maneuvers via active \ac{COM} relocation.

Full inertial steering, where internal mass movement serves as the primary directional control mechanism, has been successfully realized in spherical robots~\citep{joshiDesignAnalysisSpherical2010}, aerial platforms~\citep{hausMathematicalModellingControl2017}, and narrow single-track disks~\citep{kashkiDesignAnalysisRobust2025}.
A critical distinction in these platforms is that ground friction does not actively oppose the generated inertial wrench. 
Conversely, tracked vehicles feature extended contact patches that generate substantial turning resistance~\citep{tejiSurveyOffRoadMobile2023a}.

Yet, \citet{goodarziVehicleDynamicsControl2016} showed that on low-adherence surfaces, such as ice, inertial effects can become increasingly significant relative to wheel-ground interactions. 
To exploit this phenomenon, they investigated an inertial steering mechanism for improving the stability of a car on icy roads. 
Their approach relied on a stabilizing pendulum system, which numerical simulations showed to be effective under very low-friction conditions for which conventional tire-based stabilization becomes ineffective.
These results suggested that inertial control could provide advantages over conventional approaches in low-friction conditions. 
However, the study was limited to numerical simulations, leaving the feasibility of physical implementation experimentally unverified. 
Furthermore, inertial control was considered only as a supplementary steering mechanism, with the front wheels still generating the majority of the turning force. 
Consequently, the feasibility of using inertial steering as a primary steering mechanism for tracked ground robots on low-friction surfaces remains an open question.

\section{Theory}
\subsection{Problem definition}
In this work, the robot state comprises the body-frame twist of the \ac{COM} of the towing body, denoted $Q$, given by the longitudinal velocity, lateral velocity, and yaw rate, $\boldsymbol{\xi} = (v,w,\omega)$, together with the articulation angle, angular rate, and angular acceleration, $(\varphi,\dot{\varphi},\ddot{\varphi})$, and the longitudinal track drive speed $v_t$.
 The two input signals are the articulation and track actuator actions.
The models take the measured actuator states $\boldsymbol{u}=(v_t,\ddot{\varphi})$ as inputs rather than commands, abstracting away the low-level controllers of both actuators.
 This approach generalizes more easily to systems with different actuators and prevents a poor low-level controller model from degrading the motion model's performance. 
 
\begin{figure}[htbp]
   \centering
   \includegraphics[width=\linewidth]{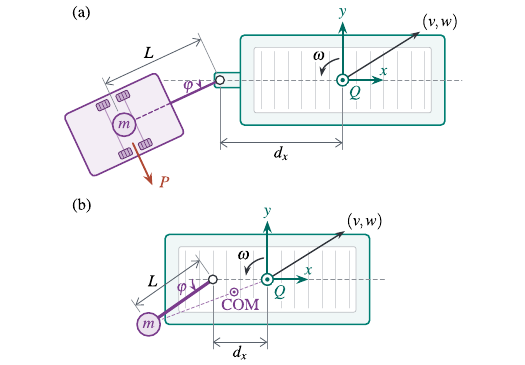} 
   \caption{Planar model of the two configurations: (a)~Articulated vehicle: the actuated hitch, at a distance~$d_x$ behind~$Q$, swings a passive wheeled trailer of mass~$m$, whose wheels receive the lateral ground force~$P$, perpendicular to the trailer. (b)~Mono-track: an actuated arm pivoting also at a distance~$d_x$ behind the main-body center of mass~$Q$ swings the mass~$m$, which shifts the \ac{COM} of the vehicle. In both cases, a mass $m$ lies at a distance~$L$ from the pivot,  and the articulation angle~$\varphi$ is measured from $-x$, positive counterclockwise. The body twist $\boldsymbol{\xi}=(v,w,\omega)$ is expressed in the body frame, with axes $x$ and $y$, attached to~$Q$. Both configurations follow the same equations of motion, except for the trailer ground force, which is zero for the mono-track, and the trailer weight, which does not load the track.}
   \label{fig:figure2}
   \vspace{-4pt}
 \end{figure}

Although an articulated vehicle and a mono-track with an internal moving mass may look like two very different dynamic systems, their multi-body dynamics are analogous.
 As shown in \cref{fig:figure2}, the articulated vehicle and the mono-track can both share the same reference frame and geometric variables.
 Indeed, the \ac{COM} of the towing body, which serves as the reference point for the reference frame, is shared between both systems.
 The mechanical arm joint of the mono-track and the articulation of the articulated vehicle are oriented the same way and are offset from $Q$ by their respective distances $d_x$ along the longitudinal axis, swinging a mass $m$ at a distance $L$ from the pivot. 
From a dynamic point of view, the only difference is that the passive trailer of an articulated vehicle interacts with the ground, while the internal moving mass of a mono-track does not. 
The articulated model is therefore supplemented with a lateral force $P$ perpendicular to its trailer.

\subsection{Track-ground contact models}
\begin{figure}[htbp]
\centering

\includegraphics[width=\linewidth]{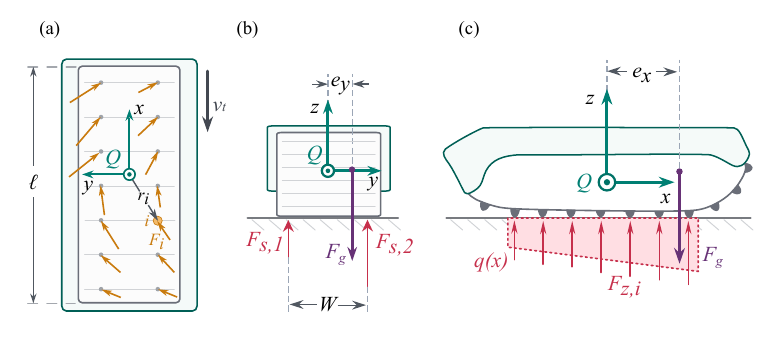}
\caption{Illustration of the discrete track contact and load-distribution model from three orthogonal views. 
\textbf{(a)} Bottom view ($x$,$y$) showing the contact points along the track and the corresponding friction forces $\boldsymbol{F}_i$, computed from $F_{z,i}$, the contact-point position $\boldsymbol{r}_i$ relative to $Q$, and the track speed $v_t$.
\textbf{(b)} Front view ($y$,$z$) showing the lateral eccentricity $e_y$ and the resulting distribution of the total normal load $F_g$ between the two contact strips, with strip reactions $F_{s,1}$ and $F_{s,2}$ of \cref{eq:stageA}.
\textbf{(c)} Side view ($x$,$z$) showing the longitudinal eccentricity $e_x$ and the resulting continuous line-load distribution $q(x)$, from which the discrete contact-point loads $F_{z,i}$ of \cref{eq:fzi} are obtained.}
\label{fig:figure3}
\vspace{-4pt}
\end{figure}

In this work, we adapted a tire-based track model designed for skid-steer configuration to a modular formulation applicable to both articulated and mono-track configurations. 
As opposed to shear-displacement-based models, tire models do not require tracking previous contact times, making them faster to compute. 
As illustrated in \cref{fig:figure3}~a), for both the mono-track and articulated configurations, we use discrete contact, where friction forces $\boldsymbol{F}_i = [F_{x,i}, F_{y,i}]^T$ are concentrated at single points $i$ on each contact pad rather than distributed over the track surface.
Discrete load distribution has been used for wheel track models~\citep{zhangSimilaritiesWheelsTracks2024} and accurately reflects how a track is supported on rigid surfaces such as ice.
The net ground wrench $\boldsymbol{T}_1$ sums the local friction forces across $N$ discretized contact points along the track.
For the contact points $i$ located at position $\boldsymbol{r}_i = [x_i, y_i]^T$ relative to the vehicle frame, $\boldsymbol{T}_1$ is expressed as
\begin{equation}
  \boldsymbol{T}_1(\varphi,v_t,\boldsymbol{\xi}) = \sum_{i=1}^{N} \begin{bmatrix} F_{x,i} \\[2pt] F_{y,i} \\[2pt] (\boldsymbol{r}_i \times \boldsymbol{F}_i)_z \end{bmatrix} = \sum_{i=1}^{N} \begin{bmatrix} F_{x,i} \\[2pt] F_{y,i} \\[2pt] x_i F_{y,i} - y_i F_{x,i} \end{bmatrix}.
  \label{eq:ground_wrench}
\end{equation}
This parameterization decouples local contact physics from global track geometry, making it suitable for arbitrary track configurations.
Because the ground is assumed rigid, track sinkage and bulldozing are neglected.

To compute the local friction forces $\boldsymbol{F}_i$, our tire model reuses the contact force calculations used by \citet{maclaurinSkidSteeringModel2011b}, with resultant friction force modeled through Pacejka's magic formula~\citep{pacejkaMagicFormulaTyre1992} as
\begin{equation}
 \lVert\boldsymbol{F}_i\rVert  = 
D_i \sin \Big(
   C \arctan \big(
        B s_{i} - E \left(
            B s_{i} - \arctan\left(
                B s_{i}
            \right)
        \right)
    \big)
\Big),
\label{eq:pacejka}
\end{equation}
where $B$, $C$, $D$ and $E$ are Pacejka's stiffness, shape, peak and curvature factor, and the local combined slip $s_i$ combines the longitudinal and lateral components of the slip at the contact point, following the definition of \citet{maclaurinSkidSteeringModel2011b}.
While $B$, $C$ and $E$ are treated as empirical parameters and fitted on subsets of data, $D$, which also represents the peak friction force, is computed according to the maximum measured friction coefficient $\mu$ and the vertical load $F_{z,i}$ carried by each contact point such that
\begin{equation}
    D_i=\mu F_{z,i}.
    \label{eq:peak}
\end{equation}
The normal load $F_{z,i}$ is obtained from a two-stage trapezoidal pressure distribution driven by an effective eccentricity ($e_x$, $e_y$) of the vehicle \ac{COM} illustrated in \cref{fig:figure3}~b) and c) and given by
\begin{equation}
  e_y=y_{cm} - \frac{h\dot{w}}{g}, \qquad
  e_x=x_{cm}  - \frac{h\dot{v}}{g}.
\end{equation}
where $x_{cm}$, $y_{cm}$ and $h$ are the position  and height of the system's \ac{COM}, while $\dot{v}$ and $\dot{w}$ are the linear accelerations and $g$ is the gravitational acceleration.
The load is first split laterally between the contact strips 
\begin{equation}\label{eq:stageA}
  F_{s,1} =  F_g\Big(\frac{1}{2}-\tfrac{e_y}{W}\Big), \qquad
  F_{s,2} =  F_g\Big(\frac{1}{2}+\tfrac{e_y}{W}\Big).
\end{equation}
with $W$ being the distance between the contact strips and $F_g$ the total weight supported by the track.
Then, following the pressure distribution formulation of \citet{wongGeneralTheorySkid2001a}, we adapt the approach to our system by defining a line-load distribution $q(x)$ along each strip of contact over the track length $\ell$ given by
\begin{equation}\label{eq:stageB}
  q_n(x) = \frac{F_{s,n}}{\ell}\left(1+\frac{12\,e_x\,x}{\ell^2}\right),
\end{equation}
where $n$ corresponds to the contact strip index.
Sampling the resulting load $q_i$ at each contact point gives the per-point normal load as 
\begin{equation}\label{eq:fzi} 
 F_{z,i} = q_i\cdot\frac{\ell}{N_s},
\end{equation} 
where $N_s$ is the number of contact points per strip.

\subsection{Articulated and mono-track model}
The classical baseline kinematic model for an articulated vehicle, assuming perfect adherence at a single contact point~\citep{corkeSteeringKinematicsCenterarticulated2001}, is given by the twist $\boldsymbol{\xi} = (v,w,\omega)$:

\begin{equation}
    v=v_t, \qquad
    w=0, \qquad
    \omega=-\frac{v_t\sin\varphi+L\dot\varphi}{d_x\cos\varphi+L}.
    \label{eq:kinematic}
\end{equation}
The design of a dynamic motion model for an articulated vehicle must take both the towing body and trailer into account.
However, modular tracks are designed to be configured with different trailer types, and modeling individual trailers is difficult, especially since they are not meant to be instrumented with wheel encoders or force sensors. 
Therefore, we abstract the trailer ground reaction wrench $\boldsymbol{T}_2$, which is set to $\boldsymbol{0}$ for a mono-track configuration, with a single-tire model based on Pacejka's magic formula.
All model components are expressed in the body frame, relative to $Q$ (\cref{fig:figure2}), and the trailer ground wrench is given by

\begin{equation}
    \boldsymbol{T}_2(\varphi,\dot{\varphi},\boldsymbol{\xi})=P(\alpha)\begin{bmatrix}
    \sin\varphi \\
    -\cos\varphi \\
    L + d_x\cos\varphi
    \end{bmatrix}.
    \label{eq:trailerwrench}
\end{equation}
Here, $P$ is computed with Pacejka's magic formula in \cref{eq:pacejka} as a function of $\alpha$, the slip angle of the trailer given by

\begin{equation}
    \alpha=\arctan \Big(\frac{-v\sin\varphi+(w-\omega d_x)\cos\varphi-L\psi}{v\cos\varphi+(w-\omega d_x)\sin\varphi}\Big)
\end{equation}
with $\psi=\omega+\dot{\varphi}$ the total angular speed of the lever.

Then, the dynamic model is derived using the Lagrangian formulation and expressed in the standard rigid-body matrix form:
 \begin{equation}
  \boldsymbol{A}(\varphi)\,\boldsymbol{\dot\xi} \;=\; \boldsymbol{T}_1(\varphi,v_t,\boldsymbol{\xi}) +\boldsymbol{T}_2(\varphi,\dot\varphi,\boldsymbol{\xi})- \boldsymbol{c}(\varphi,\boldsymbol{\xi})- \boldsymbol{b}(\varphi)\,\ddot\varphi,
  \label{eq:eom}
 \end{equation}

 where $\boldsymbol{T}_1$ comes from \cref{eq:ground_wrench} and $\boldsymbol{T}_2$ from \cref{eq:trailerwrench}, $\boldsymbol{A}(\varphi)$ is the configuration-dependent mass matrix, $\boldsymbol{b}(\varphi)\ddot\varphi$ is the reaction inertial wrench from the arm's angular acceleration, and $\boldsymbol{c}(\varphi,\boldsymbol{\xi})$ collects the velocity-dependent Coriolis and body-arm coupling terms.
 The explicit matrix formulations are given by 

\begin{align}
  \boldsymbol{A}(\varphi) &= \begin{bmatrix}
    M+m & 0 & m L_y \\
    0 & M+m & -m(d_x + L_x) \\
    m L_y & -m(d_x+L_x) & I_z + m\sigma
  \end{bmatrix}, \label{eq:A} \\[6pt]
  \boldsymbol{b}(\varphi) &= m\begin{bmatrix} L_y \\ -L_x \\ L^2+d_xL_x \end{bmatrix}, \label{eq:b} \\[6pt]
  \boldsymbol{c}(\varphi,\boldsymbol{\xi}) &= \begin{bmatrix}
    m(d_x\omega^2 + L_x\psi^2) - \omega w(M+m) \\[2pt]
    m L_y\psi^2 + \omega v(M+m) \\[2pt]
    -m (L_y d_x\dot\varphi(\psi+\omega) + \omega  (L_y w + v(L_x+d_x)))
  \end{bmatrix}, \label{eq:c}
\end{align}
where
\begin{equation}
    \sigma=L^2+d_x^2+2L_xd_x,
\end{equation}
$M$ and $I_z$ are the mass and yaw inertia of the track, $L_x=L\cos\varphi$ and $L_y=L\sin\varphi$ are the body-frame projections of the hitch length $L$.

Trajectory predictions are obtained by integrating \cref{eq:eom} forward in time from the measured state and inputs.
The measured inputs are replayed over the horizon, so both models are driven by the same sequence, and \cref{eq:eom} is integrated with the forward Euler method at a fixed step of \SI{0.05}{\second}.

\begin{figure}[htbp]
   \centering

   \includegraphics[width=1\linewidth]{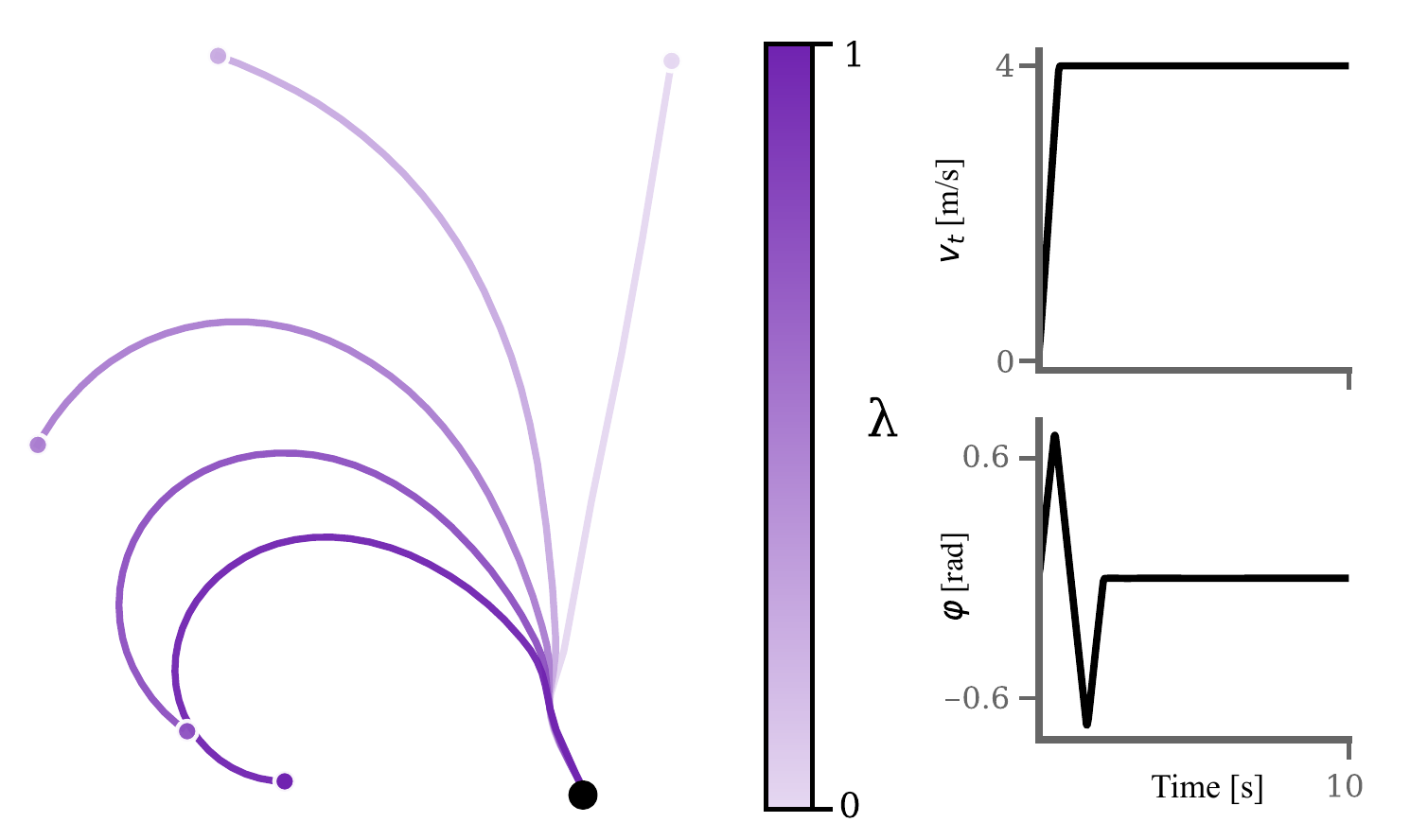}
   \caption{Simulated trajectory of \SI{10}{\second} on ice with the mixed kinematic-dynamic model predictions for different $\lambda$ values. $\lambda=0$ is equivalent to a kinematic model, while $\lambda=1$ corresponds to the predictions of our dynamic model. The trajectory represents a typical aggressive maneuver on ice where kinematic and dynamic models diverge.} 
   \label{fig:figureSweep}
\end{figure}

Thus, the articulated vehicle model has two complementary formulations: one that assumes no-slip, in \cref{eq:kinematic}, and another that accounts for inertial effects, in \cref{eq:eom}.
Following \citet{costaOnlineLearningMPC2023}, who blend the kinematic and dynamic models of a racing car through a linear weight $\lambda_k$, our mixed model predicts the state at horizon $h$ from the state and input at time $k$ as
\begin{equation}
    \boldsymbol{\xi}_{k,h}=\lambda_k\boldsymbol{f}_{dyn,h}(\boldsymbol{\xi}_k,\boldsymbol{u}_k)+(1-\lambda_k)\boldsymbol{f}_{kin,h}(\boldsymbol{\xi}_k,\boldsymbol{u}_k)
\end{equation}
where $\boldsymbol{f}_{dyn,h}$ and $\boldsymbol{f}_{kin,h}$ are the predictions of the dynamic and kinematic models at horizon $h$.
As an illustration of the phenomena at hand, \cref{fig:figureSweep} shows the effect of different values for $\lambda_k$ for the same trajectory.
When the trailer swings from one side to the other, the high deceleration that occurs at the end of its range generates a torque opposite to the curvature predicted by the classical kinematic model.
Consequently, kinematic and inertial dynamic-based models often yield diverging predictions.

\section{Methodology and Experimental setup}
\begin{figure}[htbp]
    \centering
    \includegraphics[width=1\linewidth]{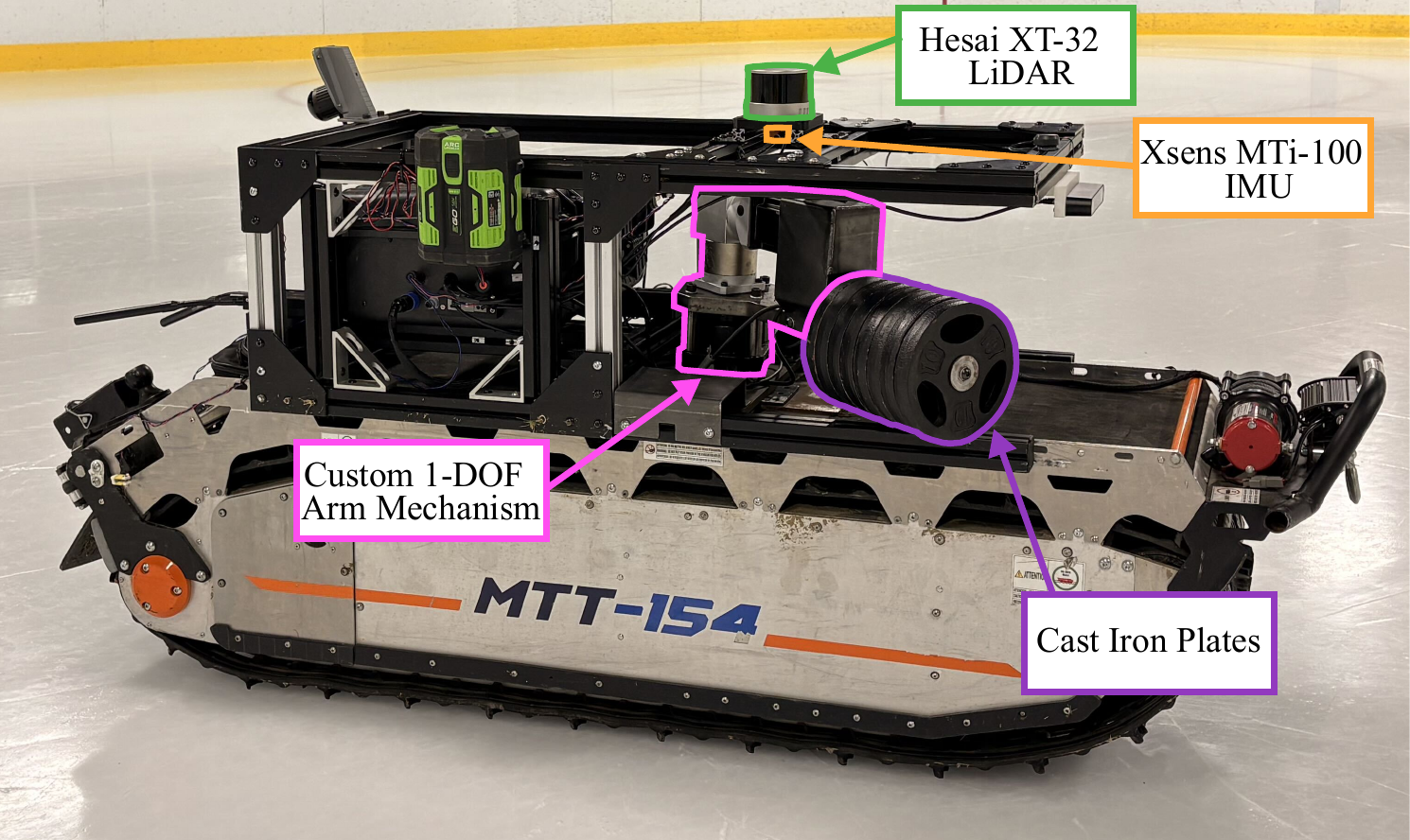}
    \caption{Experimental platform in mono-track configuration with mechanical arm installed and loaded, as well as sensors used to obtain our reference trajectories.}
    \label{fig:figure4}
\end{figure}

To test our motion model, we deployed a robotic platform on an ice rink and a grass field to compare with standard conditions. 
The platform used is called \ac{MTT-154}.
It consists of a modular, single-rubber-track vehicle with attachments and an actuated hitch. 
It is designed to enable different configurations with other MTT-154s or with passive implements such as a wheeled trailer or a winter sled. 
To evaluate the quality of our motion models, we need a reference trajectory.
To measure this reference, we equipped the platform with an Xsens MTi-100 \ac{IMU} and a Hesai XT-32 lidar as shown in \cref{fig:figure4}. 
On ice, large longitudinal and lateral slip makes track odometry unreliable, so we estimated vehicle motion from these two sensors only.
We registered each lidar scan against an incrementally built map using an \ac{ICP} mapper built on libpointmatcher~\citep{pomerleauComparingICPVariants2013}, with the \ac{IMU} providing the motion prior.
The ice surface is flat and offers few geometric features, so the point-to-plane minimization is limited to four degrees of freedom (i.e., three translations and yaw), with roll and pitch taken from the \ac{IMU}.
Position and heading are constrained by the rink boards and the surrounding structures, using points up to \SI{80}{\meter} away.
The resulting poses are available at nearly \SI{20}{\hertz}, and the reference twist $\boldsymbol{\xi}$ is obtained by differentiating consecutive poses.
Since the models are compared through the \ac{RPE} over horizons of at most \SI{5}{\second}, the reference only needs to be locally consistent over a few seconds.
Furthermore, the signal remains independent of track speed and the articulation angle, preventing track slip from biasing the evaluation metric.

Our experimental validation is twofold.
First, we attached a passive wheeled trailer to our modular track to form an articulated system configuration.
Second, to isolate and demonstrate inertial steering effects, we designed, built, and integrated a custom \num{1}-\ac{DOF} arm mechanism to move a mass along a fixed radius to emulate the inertia of a trailer without ground-reaction friction.
The arm is able to rotate at \SI{3.7}{\radian\per\second} with the equivalent of a \SI{45}{\kilogram} mass around a radius of \SI{0.40}{\meter}.
Physical parameters for both configurations are detailed in \cref{tab:parameters}, while Pacejka's parameters fitted on a subset of data and measured $\mu$ are summarized in \cref{tab:pacejka_parameters}.

\begin{table}[htbp]
\centering
\small
\setlength{\tabcolsep}{2.5pt}
\renewcommand{\arraystretch}{1.15}
\caption{Physical parameters of the platform in the mono-track (Mono.) and articulated (Art.) configurations.}
\label{tab:parameters}
\begin{tabularx}{\columnwidth}{@{}c >{\raggedright\arraybackslash}X c c c c@{}}
\toprule
 & \textbf{Characteristic} & \textbf{Symbol} & \textbf{Unit}
 & \textbf{Mono.} & \textbf{Art.} \\
\midrule
\multirow{9}{*}{\rotatebox[origin=c]{90}{\textbf{Physical}}}
 & Swung mass                   & $m$    & kg               & 45    & 156   \\
 & Pivot lever length           & $L$    & m                & 0.40  & 1.515 \\
 & Pivot--Q longitudinal offset & $d_x$  & m                & 0.308 & 1.624 \\
 & Contact patch length         & $\ell$ & m                & \multicolumn{2}{c}{1.20} \\
 & Contact points per strip     & $N_s$  & --               & \multicolumn{2}{c}{9}    \\
 & Contact points               & $N$    & --               & \multicolumn{2}{c}{18}   \\
 & Distance between strips      & $W$    & m                & \multicolumn{2}{c}{0.46} \\
 & Towing body yaw inertia      & $I_z$  & kg\,m\textsuperscript{2} & \multicolumn{2}{c}{51.1} \\
 & Towing body mass             & $M$    & kg               & \multicolumn{2}{c}{318}  \\
\bottomrule
\end{tabularx}
\end{table}

\begin{table}[htbp]
    \centering
    \small
    \setlength{\tabcolsep}{4pt}
    \renewcommand{\arraystretch}{1.15}
    \caption{Pacejka tire model parameters of the track model and trailer fitted on ice and grass. The peak factor $D$ is not fitted, it follows \cref{eq:peak}.}
    \label{tab:pacejka_parameters}
    \begin{tabular*}{\columnwidth}{@{\extracolsep{\fill}}l *{4}{S[table-format=-1.2]}@{}}
        \toprule
        & \multicolumn{2}{c}{\textbf{Ice}} & \multicolumn{2}{c}{\textbf{Grass}} \\
        \cmidrule(lr){2-3}\cmidrule(lr){4-5}
        \textbf{Parameter}
        & \multicolumn{1}{c}{\textbf{Track}} & \multicolumn{1}{c}{\textbf{Trailer}}
        & \multicolumn{1}{c}{\textbf{Track}} & \multicolumn{1}{c}{\textbf{Trailer}} \\
        \midrule
        $B$   & 6.20 & 2.10 & 1.32  & 2.80  \\
        $C$   & 1.95 & 1.92 & 1.97  & 1.61  \\
        $E$   & 1.91 & 3.27 & -1.13 & -0.30 \\
        $\mu$ & \multicolumn{2}{c}{0.1} & \multicolumn{2}{c}{0.6} \\
        \bottomrule
    \end{tabular*}
\end{table}

\section{Results}
\subsection{Impact of terrain on models}
\begin{figure}[htbp]
   \centering
   \includegraphics[width=1.0\linewidth]{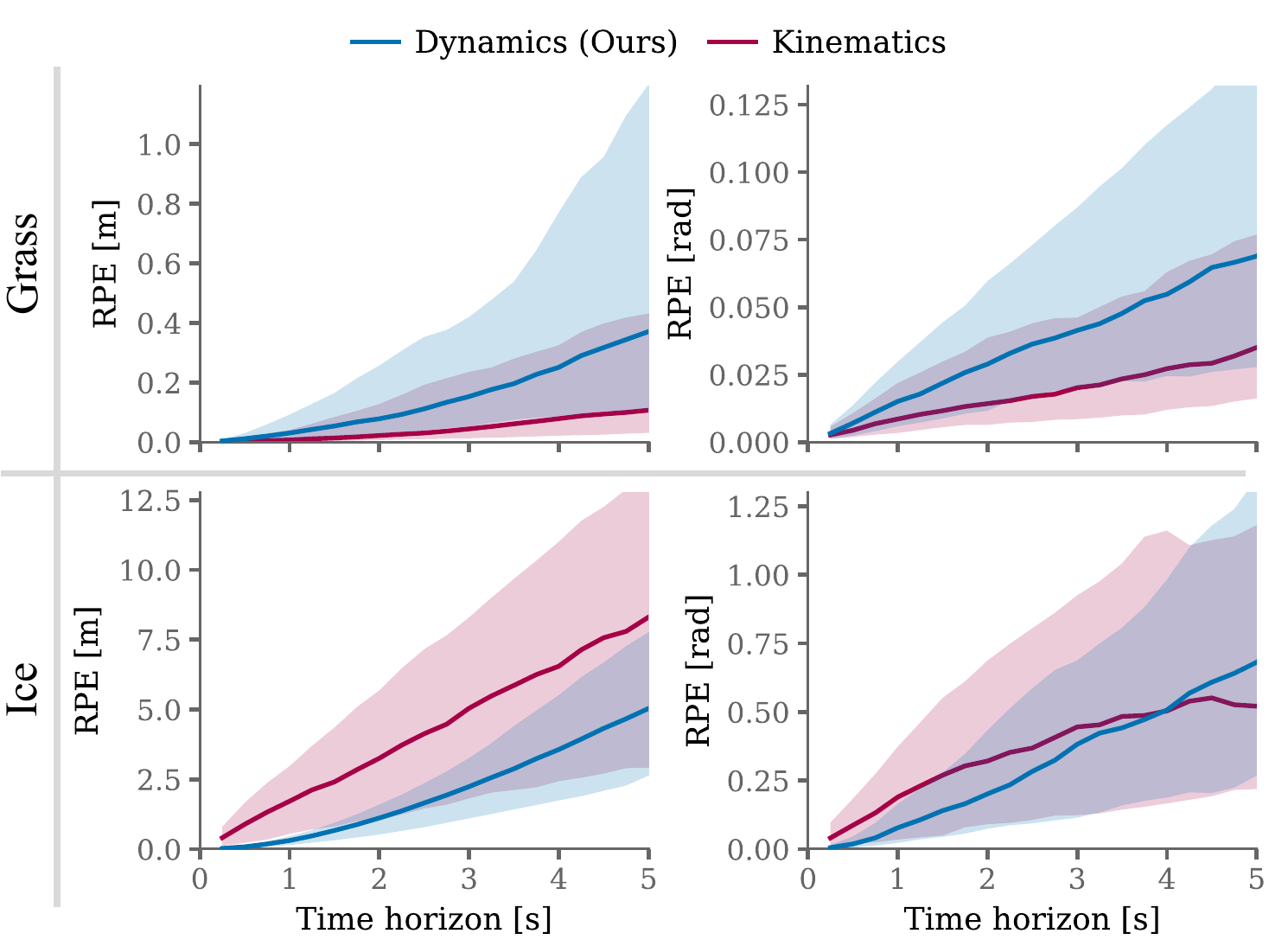}
   \caption{ \ac{RPE} in position (left column) and yaw heading (right column) of the dynamic and kinematic models over \SI{3.4}{\kilo\meter} of trajectories performed on grass and \SI{1.5}{\kilo\meter} on ice. The first row shows the results for driving on grass, while the second one shows results on ice. The full lines represent the median error, while the shaded areas represent the \ac{IQR}.
   Note that the scale of \ac{RPE} for both terrains is an order of magnitude different.}
   \label{fig:figureRPEPositionArticulationGrass}
   \vspace{-4pt}
\end{figure}

The first phase of our experiments evaluated the articulated configuration to demonstrate the impact of terrain characteristics on the predictive accuracy of kinematic and dynamic models.
To compare with standard no-slip conditions, we recorded \SI{3.4}{\kilo\meter} of driving on grass and \SI{1.5}{\kilo\meter} on ice, performed over three days of deployment.
We collected trajectories to sample the full command space by driving at speeds up to \SI{4.5}{\meter\per\second} and performing different articulation maneuvers.
\Cref{fig:figureRPEPositionArticulationGrass} presents the pose and heading prediction error of our dynamic model and the classical kinematic model over the set of trajectories collected.
The error of the dynamic model on grass is higher than that of the kinematic model at all horizons in both position and heading.
In contrast, on ice, our dynamic model had lower position error across all horizons and lower heading error up to four-second horizons.
These results confirm the literature's claim that kinematic models perform well on adherent surfaces, while the dynamic model drifted at longer prediction horizons.
On ice, conversely, the kinematic model predicts position poorly, while a dynamic model predicts it significantly better, with the kinematic median \ac{RPE} exceeding the dynamic model's \ac{IQR} upper bound.
Heading, on the other hand, is better predicted by the dynamic model on short horizons, but the trend reverses on longer horizons, where the dynamic model starts to drift and the kinematic model becomes more accurate.
These results also show that ice is more challenging than grass for any motion model to predict.
However, this comparison is not entirely direct: the limited space of an ice rink means ice trajectories include more turning maneuvers at higher average speed, and high ground friction on grass naturally damps aggressive maneuvers that instead become extreme on ice.
Nonetheless, neither model alone can predict beyond a \SI{5}{\second} horizon with \ac{RPE} below \SI{5}{\meter}, underscoring how challenging ice navigation is for motion models.

\subsection{Advantage of the mixed kinematic-dynamic model} 
\begin{figure}[htbp]
    \centering
    \includegraphics[width=1\linewidth]{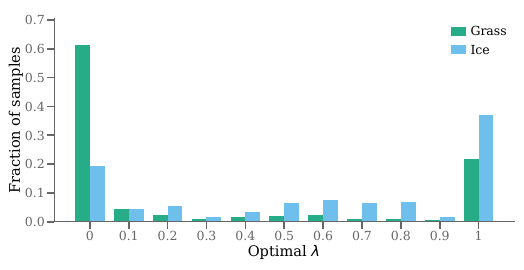}
    \caption{Share of \SI{5}{\second} windows in which each $\lambda$ gives the lowest position prediction error, over the full grass and ice articulated datasets, where $\lambda=0$ is the kinematic model alone and $\lambda=1$ the dynamic model alone. \Cref{fig:figureSweep} shows the trajectories these two ends predict.}
    \label{fig:figureBestLambda}
    \vspace{-4pt}
\end{figure}

Since neither model alone can make reliable predictions on ice, the mixed dynamic-kinematic model becomes a relevant tool.
Over a full dataset, the optimal value of $\lambda$ converges to \num{0} for grass and \num{1} for ice.
However, the optimal mixing term $\lambda$ depends not only on the driving surface, but also on the maneuver performed.
To evaluate our mixed model, we swept $\lambda$ from \num{0} to \num{1} in increments of \num{0.1} over every \SI{5}{\second} window of the grass and ice datasets.
\Cref{fig:figureBestLambda} shows the distribution of the best-performing $\lambda$ per window.
On grass, pure kinematics ($\lambda=0$) performs best in \SI{61}{\percent} of the windows, followed by pure dynamics ($\lambda=1$).
On ice, $\lambda=1$ is most frequently optimal, while $\lambda=0$ ranks second.
The distribution is bimodal and concentrated toward the extremes, suggesting a transition between inertial-dominated and adherence-dominated motion regimes.
On ice, high longitudinal and lateral slip, along with high speed and acceleration, are plausible causes for the kinematic model to behave erratically and the dynamic model to outperform it.
These results open the door to investigate the use of an adaptive $\lambda_k$ that not only adapts to the driving surface, but also responds to a metric such as slip ratio and slip angle.

\subsection{Extension to mono-track model}\label{sec:mono}
The second phase of our experiments aimed to demonstrate the inertial steering concept by removing the vehicle's standard steering component and replacing it with a mechanical arm that never contacts the ground.
Because the arm-driven mono-track is underactuated and the space to drive on an ice rink is limited, continuous driving was not possible.
Therefore, the mono-track evaluation dataset consists of short segments in which the robot was driven with a repeatable command heuristic.
The robot starts from a static state with the arm set at $\pm$\SI{85}{\degree} and is given a maximum track speed command of \SI{4.5}{\meter\per\second}. 
Once longitudinal slip exceeding \SI{50}{\percent} is detected, the arm mechanism is commanded back to position \num{0} at its maximum speed of $\pm\SI{3.7}{\radian\per\second}$. 
The track speed command is then maintained until the vehicle reaches the rink boards or a steady state.
This protocol ensures repeatability and maximizes the influence of inertial effects by providing maximum acceleration and deceleration precisely when slip is high, and therefore the friction wrench is low.

\begin{figure}[htbp]
   \centering
   \includegraphics[width=1\linewidth]{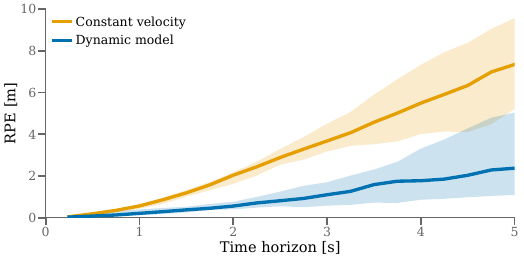}
   \caption{\ac{RPE} of the dynamic model against the constant-velocity baseline over recordings of \num{30} trajectories totaling \SI{818}{\meter} with mono-track configuration on ice performed with the experimental protocol described in \cref{sec:mono}.
   The full lines represent the median error, while the shaded areas represent the \ac{IQR}.}
   \label{fig:figureMonoRPE}
\end{figure}

To the best of our knowledge, unlike articulated systems, no prior work has proposed a kinematic or dynamic motion model for a system of the type considered in this study. 
Therefore, as a baseline, we consider a constant velocity model, a commonly used simple non-maneuvering model in motion prediction and target tracking~\citep{rongliSurveyManeuveringTarget2003}.
At the beginning of each prediction window, the predictor is initialized with the current estimated body-frame velocities $(v, w, \omega)$ from the \ac{ICP} odometry.
These quantities are then assumed constant over the prediction horizon, while the control inputs are ignored.
Because the yaw rate is maintained constant, the predicted trajectory follows a constant-curvature arc when $\omega\neq 0$, rather than a straight line.
Despite its simplicity, in trajectory prediction, the constant-velocity model has been shown to be a surprisingly strong baseline~\citep{schollerWhatConstantVelocity2020}. 
As shown in \cref{fig:figureMonoRPE}, the dynamic model yielded lower error than the constant-velocity baseline at all horizons.
These experiments demonstrate that inertial steering is predictable and relevant for track driving on low-friction surfaces.
The model reduces the error by approximately a factor of two across all the horizons.
The \ac{IQR} of the model is well below the baseline, which indicates significant performance improvement.

\begin{figure}[htbp]
   \centering
   \includegraphics[width=1\linewidth]{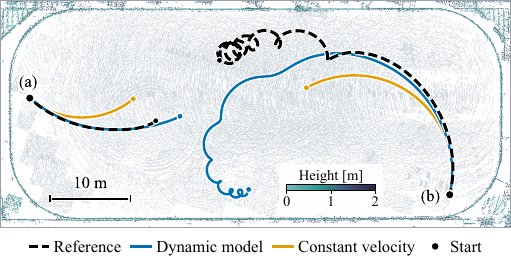}
   \caption{Mono-track trajectories on the lidar map of the ice rink, with the boards and surrounding structures coloured by height. 
   The reference (dashed) is compared with the predictions of the dynamic model and of the constant velocity baseline over the whole run. 
   \textbf{(a)} The vehicle finishes its curve, straightens its trajectory, and continues in steady forward motion. 
   \textbf{(b)} The vehicle keeps turning with high side slip and finishes in steady circular motion. The constant velocity prediction is shown over its first \SI{10}{\second}, after which it keeps circling.}
   \label{fig:figureTrajectory}
\end{figure}

The experiments show that the model reduces error on the mono-track trajectory relative to the available baseline and can also predict the end state of the trajectories.
We use \cref{fig:figureTrajectory} to present qualitative results of the models on the ice rink.
We can see examples of mono-track trajectories after an arm swing followed one of two patterns.
For the trajectory (a), the vehicle turns and then straightens out its trajectory, while for the trajectory (b) the vehicle keeps turning with high slip angle and reaches a drifting steady state.
Although the dynamic model yields larger errors in the drifting scenario, it still predicts the steady drifting behavior from the initial measured input and the trajectory duration.
These results also demonstrate that inertial steering can produce turning motion and high curvature for a vehicle without a conventional steering mechanism.
Our model's performance in predicting mono-track motion suggests these effects are predictable enough to exploit for closed-loop steering.

\section{Conclusion}
In this work, we investigated motion prediction and inertial steering of an articulated tracked vehicle operating under low-adherence conditions. 
A modular track-ground contact model was combined with a dynamic motion model and experimentally evaluated on grass and ice. 
On ice, the dynamic model outperformed the standard kinematic baseline for the articulated vehicle, demonstrating the importance of accounting for inertial effects under low-adherence conditions. 
A mixed kinematic-dynamic model further showed potential for adapting motion prediction not only to the operating surface, but also to the vehicle's maneuver. 
This suggests that a mixing parameter could be treated as an adaptive parameter within a closed-loop controller to improve tracking performance.
The experiments also demonstrated the feasibility of inertial steering for a tracked vehicle. In particular, a mono-track vehicle deprived of a conventional steering mechanism was able to generate turning motion through the controlled motion of an internal mass. 
The resulting motion could also be predicted using the proposed articulated dynamic model by representing the moving mass as a frictionless trailer, highlighting the analogous dynamics underlying both configurations.

Future work will focus on integrating the proposed motion model into an adaptive \ac{MPC} framework to investigate closed-loop control of inertial steering and its use for executing useful trajectories. 
More broadly, inertial steering may offer a promising approach to increase mobility in environments where adherence-based steering models become less effective, such as snow and sand. 
The proposed platform will therefore be deployed in the field to evaluate the approach under real, unstructured, low-adherence conditions.

\section*{Acknowledgement}
This work was supported by the  Natural  Sciences and Engineering  Research  Council of  Canada  (NSERC)  through grants ALLRP 585289-23 and CGRS M 1882877 and by the Canada Research Chair on Field Robotics (CRC-2023-00396).

Generative AI (Claude, Anthropic) was used to assist in generating the plotting code for the data-driven result figures; the underlying data, analysis, and figure content were recorded, produced, and verified by the authors.

\printbibliography

@book{wong2022theory,
author    = {Wong, J. Y.},
title     = {Theory of Ground Vehicles},
edition   = {5},
year      = {2022},
publisher = {John Wiley \& Sons},
doi       = {10.1002/9781119719984},
isbn      = {9781119719700}
}

@article{alhimdaniSteeringAnalysisArticulated1982,
  title = {Steering Analysis of Articulated Tracked Vehicles},
  author = {Alhimdani, Fahim F.},
  year = 1982,
  month = sep,
  journal = {Journal of Terramechanics},
  volume = {19},
  number = {3},
  pages = {195--209},
  issn = {0022-4898},
  doi = {10.1016/0022-4898(82)90004-0},
  urldate = {2026-09-05}
}

@article{balakrishnaModelingSlipWheeled1995,
  title = {Modeling of Slip for Wheeled Mobile Robots},
  author = {Balakrishna, R. and Ghosal, A.},
  year = 1995,
  month = feb,
  journal = {IEEE Transactions on Robotics and Automation},
  volume = {11},
  number = {1},
  pages = {126--132},
  issn = {2374-958X},
  doi = {10.1109/70.345944},
  urldate = {2026-08-21}
}

@article{bruzzoneTrackedLocomotionSystems2022,
  title = {Tracked {{Locomotion Systems}} for {{Ground Mobile Robots}}: {{A Review}}},
  shorttitle = {Tracked {{Locomotion Systems}} for {{Ground Mobile Robots}}},
  author = {Bruzzone, Luca and Nodehi, Shahab Edin and Fanghella, Pietro},
  year = 2022,
  month = aug,
  journal = {Machines},
  volume = {10},
  number = {8},
  pages = {648},
  publisher = {Multidisciplinary Digital Publishing Institute},
  issn = {2075-1702},
  doi = {10.3390/machines10080648},
  urldate = {2026-08-10},
  copyright = {http://creativecommons.org/licenses/by/3.0/},
  langid = {english}
}

@inproceedings{cariouPathFollowingVehicletrailer2010,
  title = {Path Following of a Vehicle-Trailer System in Presence of Sliding: {{Application}} to Automatic Guidance of a Towed Agricultural Implement},
  shorttitle = {Path Following of a Vehicle-Trailer System in Presence of Sliding},
  booktitle = {2010 {{IEEE}}/{{RSJ International Conference}} on {{Intelligent Robots}} and {{Systems}}},
  author = {Cariou, Christophe and Lenain, Roland and Thuilot, Benoit and Martinet, Philippe},
  year = 2010,
  month = oct,
  pages = {4976--4981},
  issn = {2153-0866},
  doi = {10.1109/IROS.2010.5652673},
  urldate = {2026-09-05}
}

@article{corkeSteeringKinematicsCenterarticulated2001,
  title = {Steering Kinematics for a Center-Articulated Mobile Robot},
  author = {Corke, P.I. and Ridley, P.},
  year = 2001,
  month = apr,
  journal = {IEEE Transactions on Robotics and Automation},
  volume = {17},
  number = {2},
  pages = {215--218},
  issn = {2374-958X},
  doi = {10.1109/70.928568},
  urldate = {2026-09-05}
}

@article{costaOnlineLearningMPC2023,
  title = {Online Learning of {{MPC}} for Autonomous Racing},
  author = {Costa, Gabriel and Pinho, Jo{\~a}o and Botto, Miguel Ayala and Lima, Pedro U.},
  year = 2023,
  month = sep,
  journal = {Robotics and Autonomous Systems},
  volume = {167},
  pages = {104469},
  issn = {0921-8890},
  doi = {10.1016/j.robot.2023.104469},
  urldate = {2026-08-21}
}

@article{febbraroObjectorientedModellingTracked2025,
  title = {Object-Oriented Modelling of a Tracked Vehicle for Agricultural Applications},
  author = {Febbraro, Alessandro and Dutto, Mattia and Bascetta, Luca and Ferretti, Gianni},
  year = 2025,
  month = mar,
  journal = {Computers and Electronics in Agriculture},
  volume = {230},
  pages = {109921},
  issn = {0168-1699},
  doi = {10.1016/j.compag.2025.109921},
  urldate = {2026-09-05}
}

@article{gonzalezSlippageEstimationCompensation2018,
  title = {Slippage Estimation and Compensation for Planetary Exploration Rovers. {{State}} of the Art and Future Challenges},
  author = {Gonzalez, Ramon and Iagnemma, Karl},
  year = 2018,
  journal = {Journal of Field Robotics},
  volume = {35},
  number = {4},
  pages = {564--577},
  issn = {1556-4967},
  doi = {10.1002/rob.21761},
  urldate = {2026-09-04},
  langid = {english}
}

@article{goodarziVehicleDynamicsControl2016,
  title = {Vehicle Dynamics Control by Using a Three-Dimensional Stabilizer Pendulum System},
  author = {Goodarzi, A. and Naghibian, M. and Choodan, D. and Khajepour, A.},
  year = 2016,
  month = dec,
  journal = {Vehicle System Dynamics},
  volume = {54},
  number = {12},
  pages = {1671--1687},
  publisher = {Taylor \& Francis},
  issn = {0042-3114},
  doi = {10.1080/00423114.2016.1228987},
  urldate = {2026-07-26}
}

@article{gustafssonSlipbasedTireroadFriction1997,
  title = {Slip-Based Tire-Road Friction Estimation},
  author = {Gustafsson, Fredrik},
  year = 1997,
  month = jun,
  journal = {Automatica},
  volume = {33},
  number = {6},
  pages = {1087--1099},
  issn = {0005-1098},
  doi = {10.1016/S0005-1098(97)00003-4},
  urldate = {2026-08-28}
}

@article{hausMathematicalModellingControl2017,
  title = {Mathematical {{Modelling}} and {{Control}} of an {{Unmanned Aerial Vehicle}} with {{Moving Mass Control Concept}}},
  author = {Haus, Tomislav and Orsag, Matko and Bogdan, Stjepan},
  year = 2017,
  month = dec,
  journal = {Journal of Intelligent \& Robotic Systems},
  volume = {88},
  number = {2},
  pages = {219--246},
  issn = {1573-0409},
  doi = {10.1007/s10846-017-0545-2},
  urldate = {2026-03-09},
  langid = {english}
}

@article{hongVerificationWheeledMobile1999,
  title = {Verification of a {{Wheeled Mobile Robot Dynamic Model}} and {{Control Ramifications}}},
  author = {Hong, Daehie and Velinsky, Steven A. and Feng, Xin},
  year = 1999,
  month = mar,
  journal = {Journal of Dynamic Systems, Measurement, and Control},
  volume = {121},
  number = {1},
  pages = {58--63},
  issn = {0022-0434},
  doi = {10.1115/1.2802442},
  urldate = {2026-08-21}
}

@article{iagnemmaControlRoboticVehicles2003,
  title = {Control of {{Robotic Vehicles}} with {{Actively Articulated Suspensions}} in {{Rough Terrain}}},
  author = {Iagnemma, Karl and Rzepniewski, Adam and Dubowsky, Steven and Schenker, Paul},
  year = 2003,
  month = jan,
  journal = {Autonomous Robots},
  volume = {14},
  number = {1},
  pages = {5--16},
  issn = {1573-7527},
  doi = {10.1023/A:1020962718637},
  urldate = {2026-01-15},
  langid = {english}
}

@article{islamClosedloopDynamicSimulationbased2012,
  title = {A Closed-Loop Dynamic Simulation-Based Design Method for Articulated Heavy Vehicles with Active Trailer Steering Systems},
  author = {Islam, Md. Manjurul and Ding, Xuejun and He, Yuping},
  year = 2012,
  month = may,
  journal = {Vehicle System Dynamics},
  volume = {50},
  number = {5},
  pages = {675--697},
  publisher = {Taylor \& Francis},
  issn = {0042-3114},
  doi = {10.1080/00423114.2011.622904},
  urldate = {2026-09-05}
}

@article{joshiDesignAnalysisSpherical2010,
  title = {Design and Analysis of a Spherical Mobile Robot},
  author = {Joshi, Vrunda A. and Banavar, Ravi N. and Hippalgaonkar, Rohit},
  year = 2010,
  month = feb,
  journal = {Mechanism and Machine Theory},
  volume = {45},
  number = {2},
  pages = {130--136},
  issn = {0094-114X},
  doi = {10.1016/j.mechmachtheory.2009.04.003},
  urldate = {2026-09-07}
}

@article{kashkiDesignAnalysisRobust2025,
  title = {Design, Analysis, and Robust Control of Inertially Actuated Two Wheeled Inverted Pendulum},
  author = {Kashki, Mohammad and Asadi, Farshid and Hurmuzlu, Yildirim},
  year = 2025,
  month = aug,
  journal = {Nonlinear Dynamics},
  volume = {113},
  number = {16},
  pages = {21215--21231},
  issn = {1573-269X},
  doi = {10.1007/s11071-025-11297-8},
  urldate = {2026-08-31},
  langid = {english}
}

@article{hac2008stability,
  author  = {Hac, Aleksander and Fulk, Daniel and Chen, Hsien},
  title   = {Stability and Control Considerations of Vehicle-Trailer Combination},
  journal = {SAE International Journal of Passenger Cars -- Mechanical Systems},
  volume  = {1},
  number  = {1},
  pages   = {925--937},
  year    = {2008},
  doi     = {10.4271/2008-01-1228},
  note    = {SAE Technical Paper 2008-01-1228}
}

@article{maclaurinSkidSteeringModel2011b,
  title = {A Skid Steering Model Using the {{Magic Formula}}},
  author = {Maclaurin, Bruce},
  year = 2011,
  month = aug,
  journal = {Journal of Terramechanics},
  volume = {48},
  number = {4},
  pages = {247--263},
  issn = {0022-4898},
  doi = {10.1016/j.jterra.2011.04.002},
  urldate = {2026-08-10}
}

@article{mahalingamNovelAlternateMultibody2021,
  title = {A Novel Alternate Multibody Model for the Longitudinal and Ride Dynamics of a Tracked Vehicle},
  author = {Mahalingam, Ilango and Padmanabhan, Chandramouli},
  year = 2021,
  month = mar,
  journal = {Vehicle System Dynamics},
  volume = {59},
  number = {3},
  pages = {433--457},
  publisher = {Taylor \& Francis},
  issn = {0042-3114},
  doi = {10.1080/00423114.2019.1693048},
  urldate = {2026-09-05}
}

@article{milicevicTheoreticalModelHighspeed2023,
  title = {Theoretical Model of High-Speed Tracked Vehicle General Motion},
  author = {Mili{\'c}evi{\'c}, Stefan V. and Blagojevi{\'c}, Ivan A.},
  year = 2023,
  journal = {FME Transactions},
  volume = {51},
  number = {3},
  pages = {449--456},
  issn = {1451-2092},
  doi = {10.5937/fme2303449M},
  urldate = {2026-09-05},
  pmid = {1451-20922303449M}
}

@article{mollKinematicDynamicModeling2025,
  title = {Kinematic and {{Dynamic Modeling}} of a {{Generalized Tractor-Trailer System}} with {{Steering Actuation}}},
  author = {Moll, Marcel and Oksanen, Timo},
  year = 2025,
  month = jan,
  journal = {IFAC-PapersOnLine},
  series = {11th {{Vienna International Conference}} on {{Mathematical Modelling MATHMOD}} 2025},
  volume = {59},
  number = {1},
  pages = {199--204},
  issn = {2405-8963},
  doi = {10.1016/j.ifacol.2025.03.035},
  urldate = {2026-09-05}
}

@inproceedings{nakamuraWheeledRobotMovable2007,
  title = {Wheeled Robot with Movable Center of Mass for Traversing over Rough Terrain},
  booktitle = {2007 {{IEEE}}/{{RSJ International Conference}} on {{Intelligent Robots}} and {{Systems}}},
  author = {Nakamura, Sousuke and Faragalli, Michele and Mizukami, Noriaki and Nakatani, Ichiro and Kunii, Yasuharu and Kubota, Takashi},
  year = 2007,
  month = oct,
  pages = {1228--1233},
  issn = {2153-0866},
  doi = {10.1109/IROS.2007.4399168},
  urldate = {2026-03-09}
}

@article{ozdemirNewContactSlip2017,
  title = {A New Contact \& Slip Model for Tracked Vehicle Transient Dynamics on Hard Ground},
  author = {{\"O}zdemir, Mehmet Nuri and K{\i}l{\i}{\c c}, Varl{\i}k and {\"U}nl{\"u}soy, Y. Samim},
  year = 2017,
  month = oct,
  journal = {Journal of Terramechanics},
  series = {Manned/{{Unmanned Ground Vehicles}}: {{Off-Road Dynamics}} and {{Mobility}}},
  volume = {73},
  pages = {3--23},
  issn = {0022-4898},
  doi = {10.1016/j.jterra.2017.07.001},
  urldate = {2026-07-13}
}

@article{pacejkaMagicFormulaTyre1992,
  title = {The {{Magic Formula Tyre Model}}},
  author = {Pacejka, Hans B. and Bakker, Egbert},
  year = 1992,
  month = jan,
  journal = {Vehicle System Dynamics},
  volume = {21},
  number = {sup001},
  pages = {1--18},
  publisher = {Taylor \& Francis},
  issn = {0042-3114},
  doi = {10.1080/00423119208969994},
  urldate = {2026-08-21}
}

@article{percyNumericalControlAlgorithm2012,
  title = {A Numerical Control Algorithm for a {{B-double}} Truck--Trailer with Steerable Trailer Wheels and Active Hitch Angles},
  author = {Percy, A and Spark, I},
  year = 2012,
  month = mar,
  journal = {Proceedings of the Institution of Mechanical Engineers, Part D: Journal of Automobile Engineering},
  volume = {226},
  number = {3},
  pages = {289--300},
  publisher = {IMECHE},
  issn = {0954-4070},
  doi = {10.1177/0954407011417355},
  urldate = {2026-09-05},
  langid = {english}
}

@article{pomerleauComparingICPVariants2013,
  title = {Comparing {{ICP}} Variants on Real-World Data Sets: {{Open-source}} Library and Experimental Protocol},
  author = {Pomerleau, Fran{\c c}ois and Colas, Francis and Siegwart, Roland and Magnenat, St{\'e}phane},
  year = 2013,
  month = feb,
  journal = {Autonomous Robots},
  volume = {34},
  number = {3},
  pages = {133--148},
  publisher = {Springer},
  issn = {1573-7527},
  doi = {10.1007/s10514-013-9327-2}
}

@article{rongliSurveyManeuveringTarget2003,
  title = {Survey of Maneuvering Target Tracking. {{Part I}}. {{Dynamic}} Models},
  author = {Rong Li, X. and Jilkov, V.P.},
  year = 2003,
  month = oct,
  journal = {IEEE Transactions on Aerospace and Electronic Systems},
  volume = {39},
  number = {4},
  pages = {1333--1364},
  issn = {1557-9603},
  doi = {10.1109/TAES.2003.1261132},
  urldate = {2026-08-31}
}

@article{schollerWhatConstantVelocity2020,
  title = {What the {{Constant Velocity Model Can Teach Us About Pedestrian Motion Prediction}}},
  author = {Sch{\"o}ller, Christoph and Aravantinos, Vincent and Lay, Florian and Knoll, Alois},
  year = 2020,
  month = apr,
  journal = {IEEE Robotics and Automation Letters},
  volume = {5},
  number = {2},
  pages = {1696--1703},
  issn = {2377-3766},
  doi = {10.1109/LRA.2020.2969925},
  urldate = {2026-08-31}
}

@article{singhIntelligentTireBased2013,
  title = {An {{Intelligent Tire Based Tire-Road Friction Estimation Technique}} and {{Adaptive Wheel Slip Controller}} for {{Antilock Brake System}}},
  author = {Singh, Kanwar B. and Ali Arat, Mustafa and Taheri, Saied},
  year = 2013,
  month = feb,
  journal = {Journal of Dynamic Systems, Measurement, and Control},
  volume = {135},
  number = {031002},
  issn = {0022-0434},
  doi = {10.1115/1.4007704},
  urldate = {2026-09-03}
}

@article{tejiSurveyOffRoadMobile2023a,
  title = {A {{Survey}} of {{Off-Road Mobile Robots}}: {{Slippage Estimation}}, {{Robot Control}}, and {{Sensing Technology}}},
  shorttitle = {A {{Survey}} of {{Off-Road Mobile Robots}}},
  author = {Teji, Mulugeta Debebe and Zou, Ting and Zeleke, Dinku Seyoum},
  year = 2023,
  month = oct,
  journal = {Journal of Intelligent \& Robotic Systems},
  volume = {109},
  number = {2},
  pages = {38},
  issn = {1573-0409},
  doi = {10.1007/s10846-023-01968-2},
  urldate = {2026-08-10},
  langid = {english}
}

@inproceedings{totaArticulatedSteeringControl2021,
  title = {Articulated {{Steering Control}} for an {{All-Terrain Tracked Vehicle}}},
  booktitle = {Advances in {{Italian Mechanism Science}}},
  author = {Tota, Antonio and Galvagno, Enrico and Velardocchia, Mauro and Rota, Emanuele and Novara, Andrea},
  editor = {Niola, Vincenzo and Gasparetto, Alessandro},
  year = 2021,
  pages = {823--830},
  publisher = {Springer International Publishing},
  address = {Cham},
  doi = {10.1007/978-3-030-55807-9_91},
  isbn = {978-3-030-55807-9},
  langid = {english}
}

@article{wangInvestigationTrackedMining2024,
  title = {Investigation of Tracked Mining Vehicles' Driving Performance at Varying Travel Speeds by Experimental and Numerical Simulation},
  author = {Wang, Ling and Chen, Xuguang and Li, Zhigang and Liu, Xixi and Liu, Xuelin and Zhang, Xian},
  year = 2024,
  month = nov,
  journal = {Applied Ocean Research},
  volume = {152},
  pages = {104142},
  issn = {0141-1187},
  doi = {10.1016/j.apor.2024.104142},
  urldate = {2026-09-05}
}

@article{wardDynamicModelBasedWheelSlip2008,
  title = {A {{Dynamic-Model-Based Wheel Slip Detector}} for {{Mobile Robots}} on {{Outdoor Terrain}}},
  author = {Ward, C.C. and Iagnemma, K.},
  year = 2008,
  month = aug,
  journal = {IEEE Transactions on Robotics},
  volume = {24},
  number = {4},
  pages = {821--831},
  issn = {1552-3098, 1941-0468},
  doi = {10.1109/TRO.2008.924945},
  urldate = {2026-08-10},
  copyright = {https://ieeexplore.ieee.org/Xplorehelp/downloads/license-information/IEEE.html}
}

@article{wongEvaluationComputerSimulation2019,
  title = {Evaluation of the Computer Simulation Model {{NTVPM}} for Assessing Military Tracked Vehicle Cross-Country Mobility},
  author = {Wong, Jo Y and Jayakumar, Paramsothy and {Preston-Thomas}, Jon},
  year = 2019,
  month = apr,
  journal = {Proceedings of the Institution of Mechanical Engineers, Part D: Journal of Automobile Engineering},
  volume = {233},
  number = {5},
  pages = {1194--1213},
  publisher = {IMECHE},
  issn = {0954-4070},
  doi = {10.1177/0954407018765504},
  urldate = {2026-09-05},
  langid = {english}
}

@article{wongGeneralTheorySkid2001a,
  title = {A General Theory for Skid Steering of Tracked Vehicles on Firm Ground},
  author = {Wong, J Y and Chiang, C F},
  year = 2001,
  month = mar,
  journal = {Proceedings of the Institution of Mechanical Engineers, Part D: Journal of Automobile Engineering},
  volume = {215},
  number = {3},
  pages = {343--355},
  publisher = {IMECHE},
  issn = {0954-4070},
  doi = {10.1243/0954407011525683},
  urldate = {2026-08-10},
  langid = {english}
}

@article{zhangSimilaritiesWheelsTracks2024,
  title = {Similarities {{Between Wheels}} and {{Tracks}}: {{A}} ``{{Tire Model}}'' for {{Tracked Vehicles}}},
  shorttitle = {Similarities {{Between Wheels}} and {{Tracks}}},
  author = {Zhang, Ruizeng and Zhou, Wei and Liu, Haiou and Gong, Jianwei and Chen, Huiyan and Khajepour, Amir},
  year = 2024,
  month = nov,
  journal = {IEEE Transactions on Vehicular Technology},
  volume = {73},
  number = {11},
  pages = {16416--16431},
  issn = {1939-9359},
  doi = {10.1109/TVT.2024.3421285},
  urldate = {2026-09-05}
}

@inproceedings{zhangSteeringControlHighSpeed2025,
  title = {Steering {{Control}} for {{High-Speed Locomotion}} of {{Wheeled-Legged Quadrupedal Robots Based}} on {{Vertical Joint Steering}}},
  booktitle = {2025 7th {{International Conference}} on {{Robotics}}, {{Intelligent Control}} and {{Artificial Intelligence}} ({{RICAI}})},
  author = {Zhang, Yifan and Zhao, Zhicheng and Shi, Xinyuan and Xiao, Shijun and Rong, Xuewen},
  year = 2025,
  month = nov,
  pages = {146--150},
  doi = {10.1109/RICAI68060.2025.11385280},
  urldate = {2026-09-05}
}

@article{zhangTerramechanicsbasedDynamicModel2024,
  title = {A {{Terramechanics-based Dynamic Model}} for {{Motion Control}} of {{Unmanned Tracked Vehicles}}},
  author = {Zhang, Ruizeng and Zhou, Wei and Liu, Haiou and Gong, Jianwei and Chen, Huiyan and Khajepour, Amir},
  year = 2024,
  journal = {IEEE Transactions on Intelligent Vehicles},
  pages = {1--14},
  issn = {2379-8904},
  doi = {10.1109/TIV.2024.3406582},
  urldate = {2026-08-31}
}

\end{document}